\pdfoutput=1

\documentclass{sigkddExp}

\usepackage{times}
\usepackage{latexsym}
\usepackage{booktabs}
\usepackage{graphicx}
\usepackage{hyperref}
\usepackage{graphicx}

\usepackage[T1]{fontenc}

\usepackage[utf8]{inputenc}

\usepackage{microtype}

\usepackage{inconsolata}

\title{Fighting Fire with Fire: Can ChatGPT Detect AI-generated Text?}

\numberofauthors{2}

\author{
\alignauthor Amrita Bhattacharjee \\
       \affaddr{School of Computing and AI}\\
       \affaddr{Arizona State University}\\
       \email{abhatt43@asu.edu}
\alignauthor Huan Liu\\
       \affaddr{School of Computing and AI}\\
       \affaddr{Arizona State University}\\
       \email{huanliu@asu.edu}
}

\begin{document}
\maketitle
\begin{abstract}
Large language models (LLMs) such as ChatGPT are increasingly being used for various use cases, including text content generation at scale. Although detection methods for such AI-generated text exist already, we investigate ChatGPT's performance as a detector on such AI-generated text, inspired by works that use ChatGPT as a data labeler or annotator. We evaluate the zero-shot performance of ChatGPT in the task of human-written vs. AI-generated text detection, and perform experiments on publicly available datasets. 
We empirically investigate if ChatGPT is symmetrically effective in detecting AI-generated or human-written text. 
Our findings provide insight on how ChatGPT and similar LLMs may be leveraged in automated detection pipelines by simply focusing on solving a specific aspect of the problem and deriving the rest from that solution. All code and data is available at \url{https://github.com/AmritaBh/ChatGPT-as-Detector}.
\end{abstract}

\section{Introduction}
\vspace{5mm}
Recently there have been incredible advancements in large language models (LLMs) that can generate high quality human-like text, with capabilities of assisting humans on a variety of tasks as well. Larger and more expressive models are released to the public frequently, either as public-facing APIs with no access to the model parameters (such as OpenAI's ChatGPT or GPT3.5 family of models~\cite{openai2023gpt,brown2020language}) or often with fully open source access (such as LLaMA~\cite{touvron2023llama}). Alongside the numerous ways in which these LLMs can aid a human user, act as an assistant and thereby improve productivity, these models can also be misused by actors with malicious intent. For example, malicious actors may use LLMs to generate misinformation and misleading content at scale and publish such content online~\cite{newsguard2023}, create fake websites for ad revenue fraud~\cite{fraud2023ads}, etc. Apart from these malicious use cases, inexperienced users may overestimate the capabilities of these LLMs. The fact that ChatGPT can confidently spew factually incorrect information, yet be fluent and cohesive in the syntax and grammar~\cite{confident2023chatgpt}, can fool newer users and mislead them. Users may use ChatGPT to write essays or reports, expecting factuality but then be penalized when flaws are evident. Especially problematic is when people use these models for high-stakes tasks, without realizing the shortcomings of these models, and eventually face dire consequences~\cite{lawyer2023chatgpt}. Given the accessibility and ease of use of such models, more and more people are using these models in their daily life, perhaps without realizing the nature of the text that is generated, often mistaking fluency and confidence as truthfulness. Therefore, in this work, we focus on the task of distinguishing between human-written and AI-generated text. 
\begin{figure}
    \centering
    \includegraphics[width=\columnwidth]{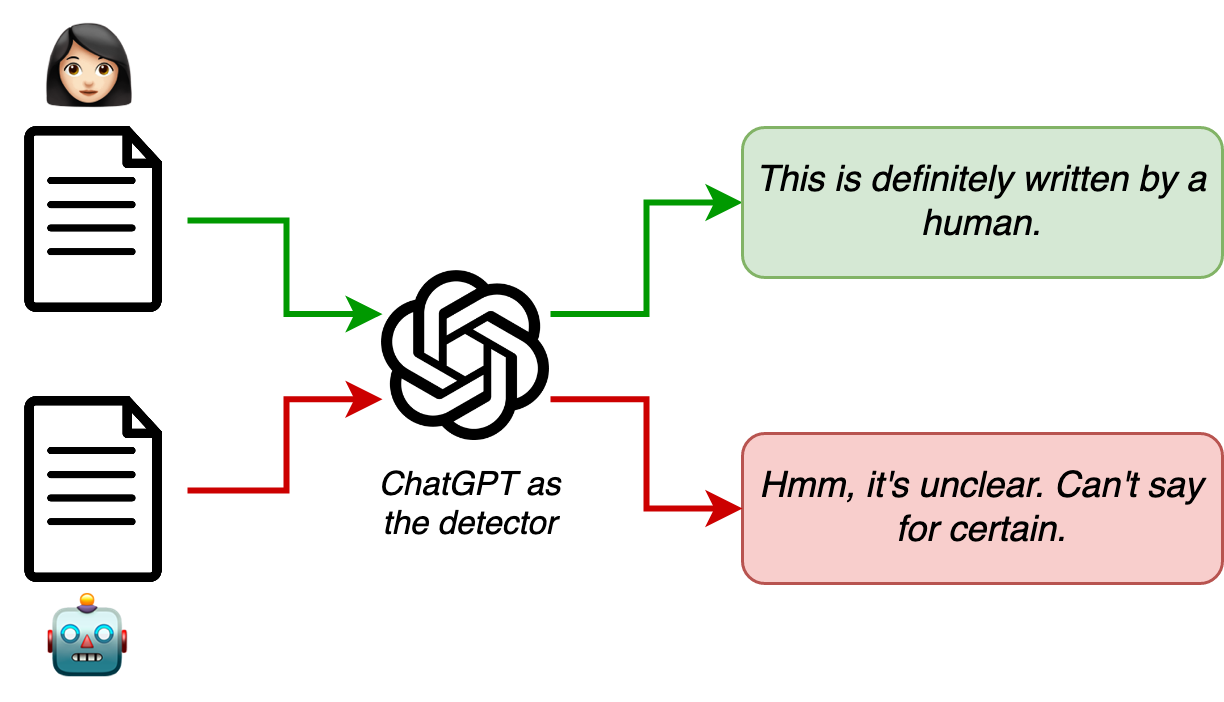}
    \caption{We use OpenAI's ChatGPT as a detector to distinguish between human-written and AI-generated text.}
    \label{fig:enter-label}
\end{figure}

Detection of AI-generated text is very challenging~\cite{ai2023detect}, with recent work~\cite{sadasivan2023can} demonstrating that this challenge is only going to get exacerbated with newer, larger, more capable LLMs. There are existing detection methods, such as feature-based classifiers~\cite{ippolito2019automatic}, methods that differentiate human and AI text using statistical measures~\cite{gehrmann2019gltr,mitchell2023detectgpt}, along with methods that use fine-tuned language models to classify text as human or AI~\cite{zellers2019defending,solaiman2019release}. In this work, however, we want to investigate the capability of ChatGPT (GPT3.5, and the more advanced GPT-4) to differentiate between human and AI-generated text. ChatGPT has been shown to perform well on a variety of NLP and NLU tasks~\cite{qin2023chatgpt}, with newer versions of the models being able to use human prompts better than older ones. In order to probe the performance of ChatGPT on detecting AI-generated text, we propose to investigate the following research question:

\textit{Can ChatGPT identify AI-generated text from a variety of generators?}


The rest of the paper is organized as follows: Section \ref{sec:background} provides a brief overview of LLMs and how they work, Section \ref{sec:chat} elaborates on our experiments and experimental settings. Section \ref{sec:results} presents and discusses our results. Section \ref{sec:related} grounds our work in the context of related work, and finally, Section \ref{sec:conclusion} concludes with a discussion of future directions.

\section{Background: Large Language \\ Models}
\label{sec:background}
\vspace{5mm}

Large language models (LLMs) are deep neural networks capable of modeling natural language. Most recent LLMs are built using transformer-based architectures, and trained on massive internet-scale corpora of data. Broadly, LLMs can be of two types: (i) autoregressive (or causal) language models, and (ii) masked language models. Autoregressive LLMs (such as the GPT family of models) are trained to predict the next word or token, given the previous token sequence in a sentence. Formally, at time step $t$, the model samples a token from the distribution $p(x_t|x_1,...,x_{t-1})$, based on some pre-defined sampling strategy, and forms the text sequence one token at a time. Sampling strategies may be greedy, top-k, nucleus sampling~\cite{holtzman2019curious}, etc., with the latter two being used commonly in recent LLMs. Masked language models (such as the BERT family of models~\cite{devlin2018bert}) are trained on cloze test tasks. Given an input text sequence, $k\%$ of the tokens are masked using a special [MASK] token, and the LLM is trained to predict the tokens in these masked locations in the sequence. Hence these models are bi-directional, unlike  autoregressive LLMs. Due to the bi-directional attention mechanisms in masked language models, these are better at natural language understanding tasks, while autoregressive GPT-style models are good at natural language generation ~\cite{liao2020probabilistically}.

\section{ChatGPT as AI-text Detector}
\label{sec:chat}
\vspace{5mm}

Recent work have used and evaluated ChatGPT for a variety of natural language tasks, annotations of data, few-shot classification settings, and even reasoning and planning (see Section \ref{sec:related}). In this work, we want to investigate whether ChatGPT can be used as a detector to identify AI-generated text. ChatGPT was trained on large amounts of text data and we would want to leverage the summarized information or understanding that ChatGPT possesses to try to identify human-written vs. AI-generated text. 

\subsubsection*{Datasets}
We use AI-generated text from the publicly available TuringBench dataset~\cite{uchendu2021turingbench}, which comprises news article style text from 19 different generators. The full list of these 19 generators are: \{GPT-1, GPT-2\_small, GPT-2\_medium, GPT-2\_large, GPT-2\_xl, GPT-2\_PyTorch, GPT-3, GROVER\_base, GROVER\_large, GROVER\_mega, CTRL, XLM, XLNET\_base, XLNET\_large, FAIR\_wmt19, FAIR\_wmt20, TRANSFORMER\_XL, PPLM\_distil, PPLM\_gpt2\}. Model sizes for each of these 19 generators in provided in Table \ref{tab:model-sizes}. 

For the human-written articles, we use the human articles from the TuringBench dataset. These are news articles from CNN and The Washington Post. 

\begin{table}[]
\centering
\begin{tabular}{@{}cc@{}}
\toprule
Model           & \# of Parameters \\ \midrule
CTRL            & 1.6B             \\
FAIR\_wmt19     & 656M             \\
FAIR\_wmt20     & 749M             \\
GPT1            & 117M             \\
GPT2\_small     & 124M             \\
GPT2\_medium    & 355M             \\
GPT2\_large     & 774M             \\
GPT2\_xl        & 1.5B             \\
GPT2\_pytorch   & 344M             \\
GPT3            & 175B             \\
GROVER\_base    & 124M             \\
GROVER\_large   & 355M             \\
GROVER\_mega    & 1.5B             \\
PPLM\_distil    & 82M              \\
PPLM\_gpt2      & 124M             \\
Transformer\_xl & 257M             \\
XLM             & 550M             \\
XLNet\_base     & 110M             \\
XLNet\_large    & 340M             \\ \bottomrule
\end{tabular}
\caption{Model sizes for the 19 generators in the TuringBench dataset.}
\label{tab:model-sizes}
\end{table}

\subsubsection*{Experimental Setting} 

We use ChatGPT (gpt-3.5-turbo model endpoint) with version as of June 13, 2023 and GPT-4 with version as of July 12, 2023 as the detector\footnote{In this paper, we use the terms `ChatGPT' to refer to the GPT-3.5 model}. We experiment with a variety of prompts (as described in the next section) and finally select an appropriate prompt for classifying each news article using ChatGPT. We set the temperature parameter to 0 to ensure minimal variability in the output since we are dealing with a classification task. For each input article, we process the text output produced by ChatGPT or GPT-4 to flag it as one of the three labels: [`human-written', `AI-generated', `unclear']. For all ChatGPT experiments, we use the test split of the datasets, which contain around $2,000$ articles. For experiments with GPT-4 as the detector we use the first $500$ samples of this test set, due to rate limit constraints on GPT-4.

\subsubsection*{Choice of Prompt}

Based on some preliminary experiments, we notice that the response from ChatGPT including `human', `AI', or `uncertain' labels for the input text depends significantly on the prompt used. For ease of experimentation and evaluation, we wanted to constrain ChatGPT's responses by using the following prompt for the input text \texttt{passage}:

\vspace{1.5mm}
\noindent\fbox{%
    \parbox{0.95\columnwidth}{%
        Task: Identify whether the given passage is generated by an AI or is human-written. Choose your answer from the given answer choices. \\
Answer choices: [``generated by AI", ``written by human", ``unsure"] \\
Passage to identify: <\texttt{passage}>
    }%
}
\vspace{1.5mm}

But we observe that with this kind of constrained prompt, ChatGPT gets confused. Not only does it fail to generate answers following the given instruction, but it also misclassifies text that it previously classified properly with a simpler prompt. It also provides incorrect labels for inputs that it was previously unsure of. 

Hence we revert back to a simpler prompt used in our preliminary experiments: 

\vspace{1.5mm}
\noindent\fbox{%
    \parbox{0.95\columnwidth}{%
        `Is the following generated by an AI or written by a human: <\texttt{text}>.'

    }%
}
\vspace{1.5mm}

where <\texttt{text}> is the main body text from a human-written or AI-generated news article from our evaluation datasets.

\begin{figure*}
    \centering
    \includegraphics[width=0.75\textwidth]{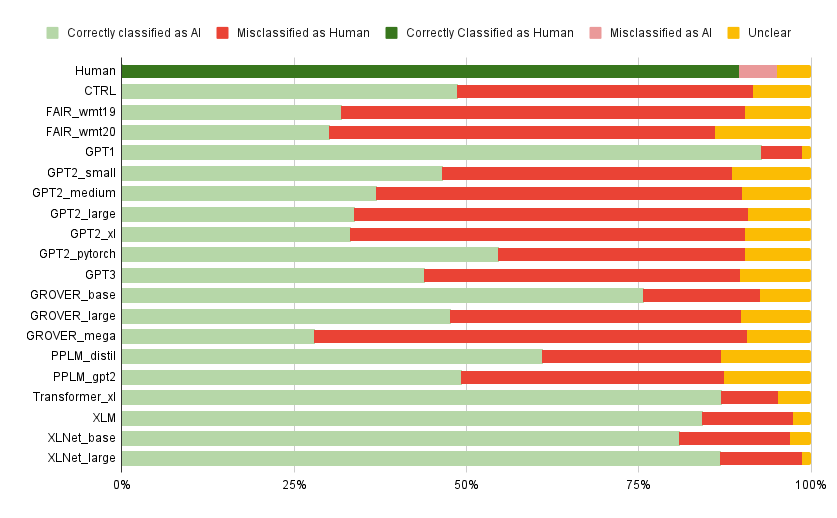}
    \caption{Performance of ChatGPT on texts generated by each of the LLMs in the TuringBench dataset, alongside performance on human-written text from TuringBench (bar at the top).}
    \label{fig:chatgpt-ai}
\end{figure*}

\section{Results and Discussion}
\label{sec:results}
\vspace{5mm}

In this section we elaborate our main experimental results with ChatGPT, with GPT-4, along with a discussion of the results, followed by some additional experiments.

\subsubsection*{ChatGPT as Detector}

We show the detection performance of ChatGPT (i.e., GPT-3.5) on text from the 19 generators in Figure \ref{fig:chatgpt-ai}. We see that for the majority of the generators, ChatGPT can identify AI-generated text less than $50\%$ of the samples, and has a very high number of false negatives, i.e., AI-generated articles misclassified as human-written. The only satisfactory performance we see is for GPT-1 with around $90\%$ samples correctly identified, and Transformer\_xl, GROVER\_base, XLM, and the XLNet models with around $75\%$ articles correctly identified as AI-generated. Therefore, our experiment demonstrates that ChatGPT is unable to identify AI-generated text across a variety of generators. We note that even if we flip the label output by ChatGPT and consider that as the final label, i.e. changing all the misclassified `human-written' labels to `AI-generated' and the correctly-classified `AI-generated' labels to now  misclassified `human-written' ones, we do not gain much in performance either, if at all. This is because the proportion of correctly classified and misclassified samples are similar for most of the generators, along with a significant fraction of samples still in the `uncertain' category.

Next, we investigate ChatGPT's performance on the human-written articles from TuringBench, and we show the distribution of the output labels in Figure \ref{fig:chatgpt-ai}. Interestingly, we see that for human-written articles, ChatGPT performs significantly better at identifying that the text is human-written, with very few samples being misclassified or even labeled as `uncertain'.

\begin{figure}
    \centering
    \includegraphics[width=\columnwidth]{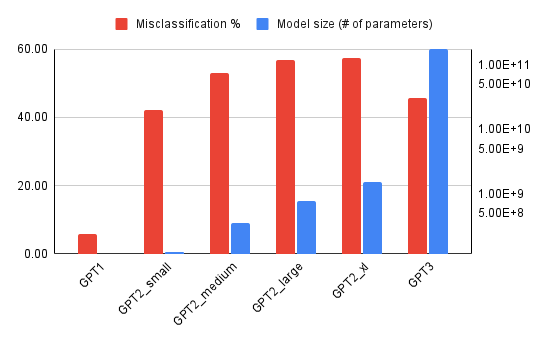}
    \caption{Percentage of AI-generated text misclassified as human-written for different model sizes of GPT. Left-axis: Misclassification $\%$; Right-axis: Model size in log-scale.}
    \label{fig:gpt-size}
\end{figure}
\begin{figure}
    \centering
    \includegraphics[width=0.7\columnwidth]{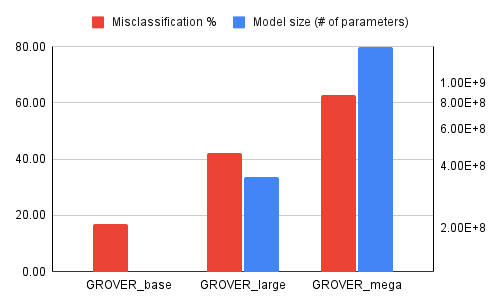}
    \caption{Percentage of AI-generated text misclassified as human-written for different model sizes of GROVER. Left-axis: Misclassification $\%$; Right-axis: Model size in log-scale.}
    \label{fig:grover-size}
\end{figure}

For the AI-generated texts, we see ChatGPT misclassifies a large fraction of these as human-written. To dig deeper into this phenomenon, we look into how the fraction of false negatives varies with the model size, within a specific model family. Figure \ref{fig:gpt-size} shows the fraction of misclassified samples with respect to the different GPT variants with GPT-1~\cite{radford2018improving} being the smallest and GPT-3~\cite{brown2020language} being the largest. We see the percentage of misclassified samples increases with an increase in model size, except for GPT-3, implying that the generation quality becomes more `human-like' with an increase in the number of model parameters. The discrepancy with GPT-3 having fewer false negatives, even though it is the largest model in our evaluation, seems to be a dataset issue since we see uncharacteristic performance on GPT-3 data even with a fully-supervised classifier. We see a similar trend for the GROVER language model~\cite{zellers2019defending} (Figure \ref{fig:grover-size}), across three of its size variants: base, large and mega. Similar to the GPT model family, we posit this behavior is due to the text quality becoming better as the model size increases and the models become more expressive. This is also consistent with the performance of other detection methods on these variants of GROVER~\cite{zellers2019defending}. 

\begin{figure*}
    \centering
    \includegraphics[width=0.75\textwidth]{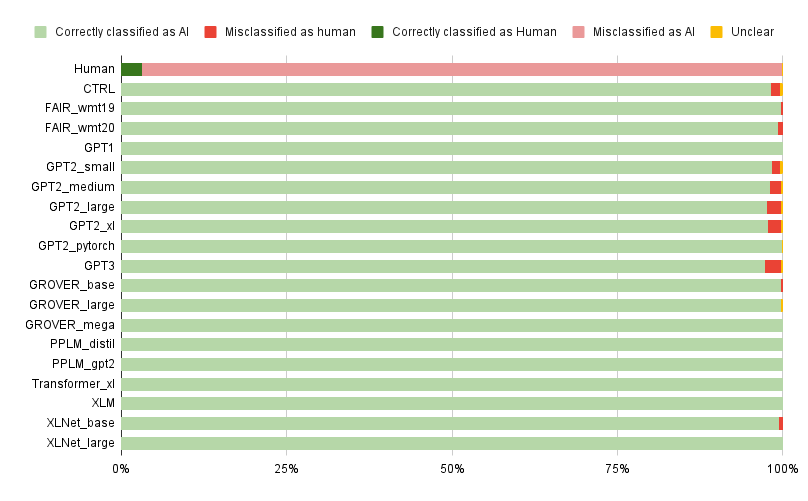}
    \caption{Performance of GPT-4 on texts generated by each of the LLMs in the TuringBench dataset, alongside performance on human-written text from TuringBench (bar at the top).}
    \label{fig:gpt4-ai}
\end{figure*}

\subsubsection*{GPT-4 as Detector}

Similarly, we show the performance of GPT-4 on the 19 generators in Figure \ref{fig:gpt4-ai}. We see that for all of the generators, GPT-4 has very good performance, correctly identifying about $97-100\%$ of all AI-generated text from the generators. Almost all the text samples are classified as AI-generated, including the human-written texts from TuringBench (top bar in Figure \ref{fig:gpt4-ai}). GPT-4 struggles to identify human-written text, and misclassifies over $~95\%$ as AI-generated. This would imply that GPT-4 is unable to differentiate between human-written and AI-generated text, for the TuringBench dataset. Interestingly, there are little to no articles (both human and AI-generated) for which GPT-4 outputs an `unclear' response.

\subsubsection*{Comparison and Discussion}

For the main experiments involving the benchmark TuringBench dataset, we see varied performance between ChatGPT and GPT-4. While there is more \textit{variability} in ChatGPT's performance, GPT-4 tends to label everything as `AI-generated'. Furthermore, we see a huge difference in the fraction of samples that ChatGPT labeled as `unclear' vs. the fraction GPT-4 labeled as `unclear'. This implies that GPT-4 is somehow more confident, even when its predictions are wrong, as in the human-written articles. Hence, these predictions are highly \textit{unreliable}. The degenerate performance of GPT-4 (i.e., labeling everything as one-class, in this case, `AI-generated') is somewhat unexpected, given the public perception that GPT-4 is better, and more capable than it's previous counterpart ChatGPT. However, recent work has shown empirical evidence that GPT-4's performance might actually be deteriorating over time~\cite{chen2023chatgpt}. This might be due to OpenAI's updates to the GPT-4 model, in order to prevent harmful generations and from people misusing the model. Similar to the drift analysis in \cite{chen2023chatgpt} we used the June 2023 version of ChatGPT and GPT-4 in our experiments, thereby revealing consistent performance degradation of GPT-4, as shown by the authors in \cite{chen2023chatgpt}.

\subsection*{Additional Experiments}

\subsubsection*{Performance on Human-written text from other sources} 

To test whether ChatGPT and GPT-4's outputs are sensitive to the specific styles, tones, topics, and other features of the human-written text from one specific dataset, we also use the human-written articles from the following datasets:

\begin{enumerate}
    \item NeuralNews~\cite{tan2020detecting}: This dataset consists of human written articles and equivalent articles generated by Grover. For our experiments, we use the human split of the dataset. These are news articles from The New York Times. 
    \item IMDb\footnote{https://huggingface.co/datasets/imdb}: This dataset comprises of movie reviews, with binary sentiment labels and is originally intended for a sentiment classification task. 
    \item TweepFake~\cite{fagni2021tweepfake}: This dataset comprises Tweets written by humans and also deepfake Tweets. For our purposes, we only use the human-written Tweets.
\end{enumerate}

\begin{figure}
    \centering
    \includegraphics[width=0.9\columnwidth]{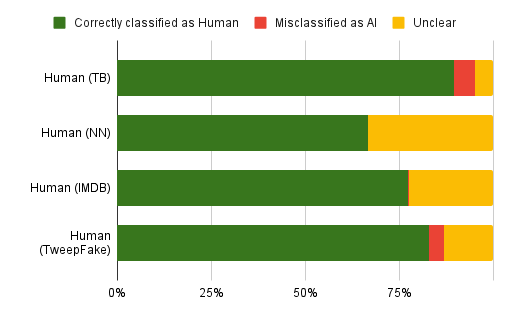}
    \caption{ChatGPT as detector: Distribution of correctly classified and misclassified samples for human-written text from four datasets: \{TuringBench, NeuralNews, IMDb, and TweepFake\}}
    \label{fig:addn}
\end{figure}

\begin{figure}
    \centering
    \includegraphics[width=0.9\columnwidth]{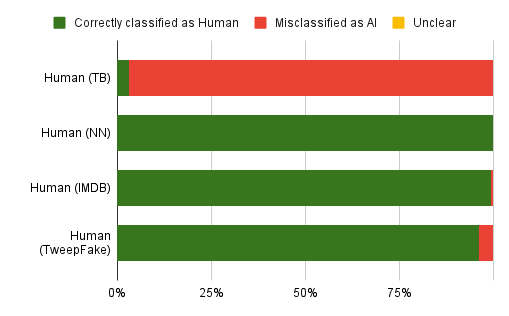}
    \caption{GPT-4 as detector: Distribution of correctly classified and misclassified samples for human-written text from from four datasets: \{TuringBench, NeuralNews, IMDb, and TweepFake\}}
    \label{fig:addn-gpt4}
\end{figure}

We show performance of ChatGPT and GPT-4 on these different tyes of human articles in Figure \ref{fig:addn} and Figure \ref{fig:addn-gpt4} respectively.
For ChatGPT as the detector (Figure \ref{fig:addn}), majority of human-written text from TuringBench, NeuralNews, IMDb and TweepFake are correctly identified as human-written. There is a significant protion of human-written text labeled as `unclear' and a much smaller fraction ($\sim4\%$ and $\sim6\%$ for TuringBench and IMDb, respectively) misclassified as AI-generated. 
Hence, we can conclude that the performance of ChatGPT in identifying human-written text is consistent across different sources and styles of human-written text data.

Interestingly, with GPT-4 as the detector (Figure \ref{fig:addn-gpt4}), we see a huge difference in performance across the four human-written datasets. While GPT-4 has almost perfect performance on human texts from NeuralNews, IMDb and very good performance on TweepFake, it misclassifies almost all human text from TuringBench. We think this might be due to dataset specific characteristics: human texts in TuringBench are relatively more `noisy' than the NeuralNews ones and tend to contain extra characters that might be artifacts from the data collection process. From our experiments, we see that GPT-4 is more sensitive to such dataset artifacts and is therefore unable to classify texts properly. This might imply that GPT-4 cannot be used reliably to identify text written by humans, unlike ChatGPT. This poor performance of recent versions of GPT-4, especially in comparison to ChatGPT, has also been reported in recent work~\cite{chen2023chatgpt}, where performance drops of even $\sim95\%$ have been observed.

\subsubsection*{Performance on ChatGPT-generated text} 

\begin{figure}
    \centering
    \includegraphics[width=0.9\columnwidth]{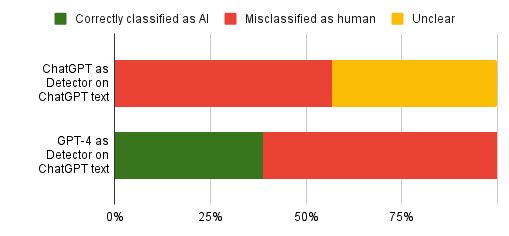}
    \caption{Performance of ChatGPT (top bar) and GPT-4 (bottom bar) on articles generated by ChatGPT.}
    \label{fig:chatgpt-articles}
\end{figure}

Since ChatGPT itself is a language model and hence has the potential of being misused, we are interested in detecting text generated by ChatGPT as well. For this, we use our own ChatGPT-generated data, created in a process similar to the TuringBench dataset~\cite{uchendu2021turingbench}. More precisely, we use the same human article sources as in ~\cite{uchendu2021turingbench}, and use the headlines of the articles to generate equivalent ChatGPT articles, using the following prompt: 

\vspace{1.5mm}
\noindent\fbox{%
    \parbox{0.95\columnwidth}{%
 Generate a news article with the headline `<headline>'.
    }%
}
\vspace{1.5mm}

 where <headline> comes from the actual human-written article. For creating this dataset, we use the ChatGPT (GPT-3.5) version as on March 14, 2023. For the rest of this paper, we refer to this dataset as ChatNews, for brevity. We use ChatGPT and GPT-4 as detectors on ChatNews, and report the performance in Figure \ref{fig:chatgpt-articles}. We see that ChatGPT misclassifies over $50\%$ of ChatNews articles as human-written, and for the remaining, ChatGPT outputs the label `uncertain'. Only 2 out of 2,000 ChatNews are correctly identified as AI-generated. This poor performance may be due to articles in ChatNews being extremely high quality and human-like, and essentially indistinguishable from actual human-written text. However, when we use GPT-4 as a detector on the same ChatNews dataset, we see more promising results. Interestingly, GPT-4 can correctly identify \textit{some} fraction of ChatNews articles (around $38\%$) while ChatGPT fails completely. This gives an insight into how newer, larger and perhaps more capable language models may potentially be used to detect text from older language models. 

\pagebreak
\section{Related Work}
\label{sec:related}
\vspace{2mm}

\subsubsection*{ChatGPT as a detector or expert} 

Recent language models ChatGPT and GPT-4 have shown impressive performance on a variety of NLP tasks~\cite{qin2023chatgpt} such as natural language inference, question-answering, sentiment analysis, named entity recognition, etc. There is also empirical evidence of GPT-4 being able to perform discriminative tasks such as identifying PII (personally identifiable information), fact-checking etc.~\cite{bubeck2023sparks}. LLMs have also been evaluated as annotators~\cite{efrat2020turking,he2023annollm} with recent work showing some LLMs perform at par or even out-perform human crowd workers for text annotation and question-answering~\cite{guo2023close,gilardi2023chatgpt,tornberg2023chatgpt}. Some recent work also demonstrates how to use an LLM as a controller to use multiple models for AI tasks. Interestingly, there is also evidence~\cite{kocon2023chatgpt} that ChatGPT may not perform well for more subjective NLP tasks.

\subsubsection*{The Landscape of AI-generated text and its detection} 

The advent of large language models(LLMs) and especially LLM assistants like ChatGPT has normalized the use of AI-generated text for a variety of purposes. Lay persons are also able to use LLMs for work, homework, leisure, or in some cases, even to mislead readers. While such tools can indeed boost productivity and inspire creative thinking, understanding the limitations of these is also important~\cite{bian2023chatgpt,alkaissi2023artificial,bang2023multitask}. Given the potential for misuse of LLMs, research on the detection of AI-generated text has gained traction. While several works have shown that humans struggle at identifying AI-generated language~\cite{jakesch2023human,clark2021all}, a variety of computational methods for detection also exist, including feature-based methods~\cite{ippolito2019automatic,kumarage2023stylometric}, methods exploiting difference in statistical measures across human and AI-generated text~\cite{gehrmann2019gltr,mitchell2023detectgpt}, more black-box type methods involving fine-tuned language models as the detector backbone~\cite{zellers2019defending,solaiman2019release}, etc. With the popularity of ChatGPT and other conversational language models, many commercial AI content detectors have also been released for use, and marketed for use-cases such as plagiarism detection\footnote{https://docs.thehive.ai/docs/ai-generated-text-detection}\footnote{https://copyleaks.com/ai-content-detector}. Prominent ones include OpenAI's detector\footnote{https://openai.com/blog/new-ai-classifier-for-indicating-ai-written-text}, the famous ZeroGPT detector\footnote{https://www.zerogpt.com/}, etc. Another recent line of research in the direction of AI-ghenerated text detection is that of watermarking~\cite{zhao2023provable,zhao2023protecting,kirchenbauer2023watermark,christ2023undetectable} whereby indistinguishable artifacts are embedded into the text, that can be identified by computational or statistical detection methods, but not by a human reader.


\section{Conclusion \& Future Work}
\label{sec:conclusion}
\vspace{5mm}

In this work, we investigated the capability of ChatGPT, a large language model, to detect AI-generated text. Our experiments demonstrate an interesting finding that even though ChatGPT struggles to identify AI-generated text, it does perform well on human-written text. This asymmetric performance of ChatGPT can be leveraged to build detectors that focus on identifying human-written text, and thus effectively solve the problem of AI-generated text detection, albeit in an indirect way. A few important takeaways from this empirical analysis would be:
\begin{itemize}
    \item ChatGPT (GPT3.5) has better, more reliable performance than GPT-4 in identifying AI-generated text vs. human-written text. 
    \item GPT-4 seems to be extremely sensitive to noise and dataset artifacts, such as those that are a result of scraping text from the internet. 
    \item GPT-4's performance is deteriorating over time, and therefore results from using GPT-4 for identifying human-written text may be unreliable and inconclusive. 
    \item The asymmetric performance of ChatGPT may be leveraged in a downstream detection task: ChatGPT (GPT3.5) may be used to specifically identify human-written texts reliably, thereby solving a portion of the AI-generated vs. human-written text detection task. 
    \item Newer, larger generators may be used to detect text from older generators, such as using GPT-4 to identify ChatGPT-generated text.
\end{itemize}

In future work, we would want to explore \textit{why} this difference in performance exists, and why ChatGPT is much better at identifying human-written text, with significantly less percentage of false negatives (i.e. human-written but misclassified as AI). One hypothesis could be that ChatGPT and these new large language models have been trained on huge corpora of text from the internet. Most of these datasets have data only till 2021, wherein much of the text on the internet was human-written (although with the pervasiveness of ChatGPT and other recent LLMs, the fraction of human to AI text on the internet would possibly change). Therefore, ChatGPT has `seen' different styles of human writing, how human language flows, and therefore has a better understanding of what human-written text would look like. This subtle capability of ChatGPT may be leveraged to build automated detection pipelines to check the probability of a text being AI-generated. Other interesting future directions for using ChatGPT (or other LLMs) for this task may include few-shot prompting based methods, and ensemble methods leveraging multiple LLMs or feature-based classifiers. 

\section*{Acknowledgments}
\vspace{2mm}
This work is supported by the DARPA SemaFor project (HR001120-C0123) and by the Office of Naval Research via grant no. N00014-21-1-4002. The views, opinions and/or findings expressed are those of the authors and should not be interpreted as representing the official views or policies of the Department of Defense or the U.S. Government.

\bibliographystyle{abbrv}
\bibliography{anthology,custom,references}

\begin{thebibliography}{10}

\bibitem{alkaissi2023artificial}
H.~Alkaissi and S.~I. McFarlane.
\newblock Artificial hallucinations in chatgpt: implications in scientific
  writing.
\newblock {\em Cureus}, 15(2), 2023.

\bibitem{confident2023chatgpt}
S.~Allardice.
\newblock The confident wrongness of chatgpt, 2023.
\newblock Accessed on June 29, 2023.

\bibitem{bang2023multitask}
Y.~Bang, S.~Cahyawijaya, N.~Lee, W.~Dai, D.~Su, B.~Wilie, H.~Lovenia, Z.~Ji,
  T.~Yu, W.~Chung, et~al.
\newblock A multitask, multilingual, multimodal evaluation of chatgpt on
  reasoning, hallucination, and interactivity.
\newblock {\em arXiv preprint arXiv:2302.04023}, 2023.

\bibitem{bian2023chatgpt}
N.~Bian, X.~Han, L.~Sun, H.~Lin, Y.~Lu, and B.~He.
\newblock Chatgpt is a knowledgeable but inexperienced solver: An investigation
  of commonsense problem in large language models.
\newblock {\em arXiv preprint arXiv:2303.16421}, 2023.

\bibitem{lawyer2023chatgpt}
M.~Bohanno.
\newblock Lawyer used chatgpt in court—and cited fake cases. a judge is
  considering sanctions, 2023.
\newblock Accessed on June 29, 2023.

\bibitem{brown2020language}
T.~Brown, B.~Mann, N.~Ryder, M.~Subbiah, J.~D. Kaplan, P.~Dhariwal,
  A.~Neelakantan, P.~Shyam, G.~Sastry, A.~Askell, et~al.
\newblock Language models are few-shot learners.
\newblock {\em Advances in neural information processing systems},
  33:1877--1901, 2020.

\bibitem{bubeck2023sparks}
S.~Bubeck, V.~Chandrasekaran, R.~Eldan, J.~Gehrke, E.~Horvitz, E.~Kamar,
  P.~Lee, Y.~T. Lee, Y.~Li, S.~Lundberg, et~al.
\newblock Sparks of artificial general intelligence: Early experiments with
  gpt-4.
\newblock {\em arXiv preprint arXiv:2303.12712}, 2023.

\bibitem{chen2023chatgpt}
L.~Chen, M.~Zaharia, and J.~Zou.
\newblock How is chatgpt's behavior changing over time?
\newblock {\em arXiv preprint arXiv:2307.09009}, 2023.

\bibitem{christ2023undetectable}
M.~Christ, S.~Gunn, and O.~Zamir.
\newblock Undetectable watermarks for language models.
\newblock {\em arXiv preprint arXiv:2306.09194}, 2023.

\bibitem{clark2021all}
E.~Clark, T.~August, S.~Serrano, N.~Haduong, S.~Gururangan, and N.~A. Smith.
\newblock All that's' human'is not gold: Evaluating human evaluation of
  generated text.
\newblock {\em arXiv preprint arXiv:2107.00061}, 2021.

\bibitem{devlin2018bert}
J.~Devlin, M.-W. Chang, K.~Lee, and K.~Toutanova.
\newblock Bert: Pre-training of deep bidirectional transformers for language
  understanding.
\newblock {\em arXiv preprint arXiv:1810.04805}, 2018.

\bibitem{efrat2020turking}
A.~Efrat and O.~Levy.
\newblock The turking test: Can language models understand instructions?
\newblock {\em arXiv preprint arXiv:2010.11982}, 2020.

\bibitem{fagni2021tweepfake}
T.~Fagni, F.~Falchi, M.~Gambini, A.~Martella, and M.~Tesconi.
\newblock Tweepfake: About detecting deepfake tweets.
\newblock {\em Plos one}, 16(5):e0251415, 2021.

\bibitem{gehrmann2019gltr}
S.~Gehrmann, H.~Strobelt, and A.~M. Rush.
\newblock Gltr: Statistical detection and visualization of generated text.
\newblock {\em arXiv preprint arXiv:1906.04043}, 2019.

\bibitem{gilardi2023chatgpt}
F.~Gilardi, M.~Alizadeh, and M.~Kubli.
\newblock Chatgpt outperforms crowd-workers for text-annotation tasks.
\newblock {\em arXiv preprint arXiv:2303.15056}, 2023.

\bibitem{guo2023close}
B.~Guo, X.~Zhang, Z.~Wang, M.~Jiang, J.~Nie, Y.~Ding, J.~Yue, and Y.~Wu.
\newblock How close is chatgpt to human experts? comparison corpus, evaluation,
  and detection.
\newblock {\em arXiv preprint arXiv:2301.07597}, 2023.

\bibitem{he2023annollm}
X.~He, Z.~Lin, Y.~Gong, A.~Jin, H.~Zhang, C.~Lin, J.~Jiao, S.~M. Yiu, N.~Duan,
  W.~Chen, et~al.
\newblock Annollm: Making large language models to be better crowdsourced
  annotators.
\newblock {\em arXiv preprint arXiv:2303.16854}, 2023.

\bibitem{ai2023detect}
M.~Heikkilä.
\newblock Why detecting ai-generated text is so difficult (and what to do about
  it), 2023.
\newblock Accessed on June 29, 2023.

\bibitem{holtzman2019curious}
A.~Holtzman, J.~Buys, L.~Du, M.~Forbes, and Y.~Choi.
\newblock The curious case of neural text degeneration.
\newblock {\em arXiv preprint arXiv:1904.09751}, 2019.

\bibitem{ippolito2019automatic}
D.~Ippolito, D.~Duckworth, C.~Callison-Burch, and D.~Eck.
\newblock Automatic detection of generated text is easiest when humans are
  fooled.
\newblock {\em arXiv preprint arXiv:1911.00650}, 2019.

\bibitem{jakesch2023human}
M.~Jakesch, J.~T. Hancock, and M.~Naaman.
\newblock Human heuristics for ai-generated language are flawed.
\newblock {\em Proceedings of the National Academy of Sciences},
  120(11):e2208839120, 2023.

\bibitem{kirchenbauer2023watermark}
J.~Kirchenbauer, J.~Geiping, Y.~Wen, J.~Katz, I.~Miers, and T.~Goldstein.
\newblock A watermark for large language models.
\newblock {\em arXiv preprint arXiv:2301.10226}, 2023.

\bibitem{kocon2023chatgpt}
J.~Koco{\'n}, I.~Cichecki, O.~Kaszyca, M.~Kochanek, D.~Szyd{\l}o, J.~Baran,
  J.~Bielaniewicz, M.~Gruza, A.~Janz, K.~Kanclerz, et~al.
\newblock Chatgpt: Jack of all trades, master of none.
\newblock {\em arXiv preprint arXiv:2302.10724}, 2023.

\bibitem{kumarage2023stylometric}
T.~Kumarage, J.~Garland, A.~Bhattacharjee, K.~Trapeznikov, S.~Ruston, and
  H.~Liu.
\newblock Stylometric detection of ai-generated text in twitter timelines.
\newblock {\em arXiv preprint arXiv:2303.03697}, 2023.

\bibitem{liao2020probabilistically}
Y.~Liao, X.~Jiang, and Q.~Liu.
\newblock Probabilistically masked language model capable of autoregressive
  generation in arbitrary word order.
\newblock {\em arXiv preprint arXiv:2004.11579}, 2020.

\bibitem{mitchell2023detectgpt}
E.~Mitchell, Y.~Lee, A.~Khazatsky, C.~D. Manning, and C.~Finn.
\newblock Detectgpt: Zero-shot machine-generated text detection using
  probability curvature.
\newblock {\em arXiv preprint arXiv:2301.11305}, 2023.

\bibitem{openai2023gpt}
R.~OpenAI.
\newblock Gpt-4 technical report.
\newblock {\em arXiv}, pages 2303--08774, 2023.

\bibitem{qin2023chatgpt}
C.~Qin, A.~Zhang, Z.~Zhang, J.~Chen, M.~Yasunaga, and D.~Yang.
\newblock Is chatgpt a general-purpose natural language processing task solver?
\newblock {\em arXiv preprint arXiv:2302.06476}, 2023.

\bibitem{radford2018improving}
A.~Radford, K.~Narasimhan, T.~Salimans, I.~Sutskever, et~al.
\newblock Improving language understanding by generative pre-training.
\newblock 2018.

\bibitem{fraud2023ads}
T.~Ryan-Mosley.
\newblock Junk websites filled with ai-generated text are pulling in money from
  programmatic ads, 2023.
\newblock Accessed on July 1, 2023.

\bibitem{sadasivan2023can}
V.~S. Sadasivan, A.~Kumar, S.~Balasubramanian, W.~Wang, and S.~Feizi.
\newblock Can ai-generated text be reliably detected?
\newblock {\em arXiv preprint arXiv:2303.11156}, 2023.

\bibitem{newsguard2023}
M.~Sadeghi and L.~Arvanitis.
\newblock Rise of the newsbots: Ai-generated news websites proliferating
  online, 2023.
\newblock Accessed on July 1, 2023.

\bibitem{solaiman2019release}
I.~Solaiman, M.~Brundage, J.~Clark, A.~Askell, A.~Herbert-Voss, J.~Wu,
  A.~Radford, G.~Krueger, J.~W. Kim, S.~Kreps, et~al.
\newblock Release strategies and the social impacts of language models.
\newblock {\em arXiv preprint arXiv:1908.09203}, 2019.

\bibitem{tan2020detecting}
R.~Tan, B.~A. Plummer, and K.~Saenko.
\newblock Detecting cross-modal inconsistency to defend against neural fake
  news.
\newblock {\em arXiv preprint arXiv:2009.07698}, 2020.

\bibitem{tornberg2023chatgpt}
P.~T{\"o}rnberg.
\newblock Chatgpt-4 outperforms experts and crowd workers in annotating
  political twitter messages with zero-shot learning.
\newblock {\em arXiv preprint arXiv:2304.06588}, 2023.

\bibitem{touvron2023llama}
H.~Touvron, T.~Lavril, G.~Izacard, X.~Martinet, M.-A. Lachaux, T.~Lacroix,
  B.~Rozi{\`e}re, N.~Goyal, E.~Hambro, F.~Azhar, et~al.
\newblock Llama: Open and efficient foundation language models.
\newblock {\em arXiv preprint arXiv:2302.13971}, 2023.

\bibitem{uchendu2021turingbench}
A.~Uchendu, Z.~Ma, T.~Le, R.~Zhang, and D.~Lee.
\newblock Turingbench: A benchmark environment for turing test in the age of
  neural text generation.
\newblock {\em arXiv preprint arXiv:2109.13296}, 2021.

\bibitem{zellers2019defending}
R.~Zellers, A.~Holtzman, H.~Rashkin, Y.~Bisk, A.~Farhadi, F.~Roesner, and
  Y.~Choi.
\newblock Defending against neural fake news.
\newblock {\em Advances in neural information processing systems}, 32, 2019.

\bibitem{zhao2023provable}
X.~Zhao, P.~Ananth, L.~Li, and Y.-X. Wang.
\newblock Provable robust watermarking for ai-generated text.
\newblock {\em arXiv preprint arXiv:2306.17439}, 2023.

\bibitem{zhao2023protecting}
X.~Zhao, Y.-X. Wang, and L.~Li.
\newblock Protecting language generation models via invisible watermarking.
\newblock {\em arXiv preprint arXiv:2302.03162}, 2023.

\end{thebibliography}

\end{document}